\documentclass[10pt,twocolumn,letterpaper]{article}

\usepackage[pagenumbers]{st} 

\definecolor{cvprblue}{rgb}{0.21,0.49,0.74}
\usepackage[pagebackref,breaklinks,colorlinks,allcolors=cvprblue]{hyperref}
\usepackage[most]{tcolorbox}
\usepackage{amsmath,mathtools,mathrsfs,amsthm}
\usepackage{multirow}

\usepackage{colortbl}
\definecolor{mygray}{gray}{.92}
\usepackage{algpseudocode}
\usepackage{booktabs}
\usepackage{algorithm}
\usepackage{multirow}
\usepackage{bm}
\usepackage{amsmath}
\newcommand{\nameshort}[1]{Ours w/o Beam}
\newcommand{\nameshortbeam}[1]{Ours w Beam}
\newcommand{\myparagraph}[1]{\noindent{\textbf{#1}\quad}}
\def\paperID{17207} 
\def\confName{CVPR}
\def\confYear{2026}

\title{Suppressing VLM Hallucinations with Spectral Representation Filtering}

\author{Ameen Ali\\
Tel Aviv University\\
{\tt\small ameenali@mail.tau.ac.il}
\and
Tamim Zoabi\\
Tel Aviv University\\
{\tt\small tamimzoabi@mail.tau.ac.il}
\and
Lior Wolf\\
Tel Aviv University\\
{\tt\small wolf@cs.tau.ac.il}
}

\begin{document}

\maketitle
\begin{abstract}
Vision-language models (VLMs) frequently produce hallucinations in the form of descriptions of objects, attributes, or relations that do not exist in the image due to over-reliance on language priors and imprecise cross-modal grounding. We introduce Spectral Representation Filtering (SRF), a lightweight, training-free method to suppress such hallucinations by analyzing and correcting the covariance structure of the model's  representations. SRF identifies low-rank hallucination modes through eigendecomposition of the covariance of the differences between features collected for truthful and hallucinatory captions, revealing structured biases in the feature space. A soft spectral filter then attenuates these modes in the feed-forward projection weights of deeper vLLM layers, equalizing feature variance while preserving semantic fidelity. Unlike decoding or retraining-based approaches, SRF operates entirely post-hoc, incurs zero inference overhead, and requires no architectural modifications. Across three families of VLMs (LLaVA-1.5, MiniGPT-4, and mPLUG-Owl2), SRF consistently reduces hallucination rates on MSCOCO, POPE-VQA, and other visual tasks benchmarks, achieving state-of-the-art faithfulness without degrading caption quality.
\end{abstract}

\section{Introduction}
Vision-language models (VLMs) have achieved remarkable success in bridging visual perception and natural language processing~\cite{liu2023visual,ye2023mplugowl2,ye2023mplugowl,chen2023minigptv2,zhu2023minigpt}, powering diverse applications such as image captioning, visual question answering (VQA), and multimodal reasoning. By integrating pre-trained vision encoders with large language models (LLMs), these systems generate fluent, context-aware descriptions of visual scenes. Despite their prowess, VLMs remain plagued by \emph{object hallucinations} (OH): the fabrication of objects, attributes, or spatial relations absent from the input image, often stemming from linguistic priors or distributional biases in training data \cite{huang2024visual,rohrbach-etal-2018-object,bai2024hallucination,zhong-etal-2024-investigating,huang-etal-2024-visual}.

Existing mitigation strategies fall into two categories: decoding-time adjustments, such as constrained beam search \cite{freitag-al-onaizan-2017-beam} or logit debiasing~\cite{chuang2024dola,leng2024mitigating}, which modulate generation to promote grounded outputs; and post-hoc refinements such as Woodpecker \cite{yin2024woodpecker}, LURE \cite{zhou2024analyzing}), and the reference-guided editing method HALC \cite{chen2024halc}. While these yield gains in controlled scenarios, they introduce substantial overhead, often 5--10x inference slowdowns, demand task-specific adaptations, or rely on external resources.

\begin{figure}[t]
    \centering
     \hfill
    \includegraphics[width=1\linewidth,height=0.4\textwidth]{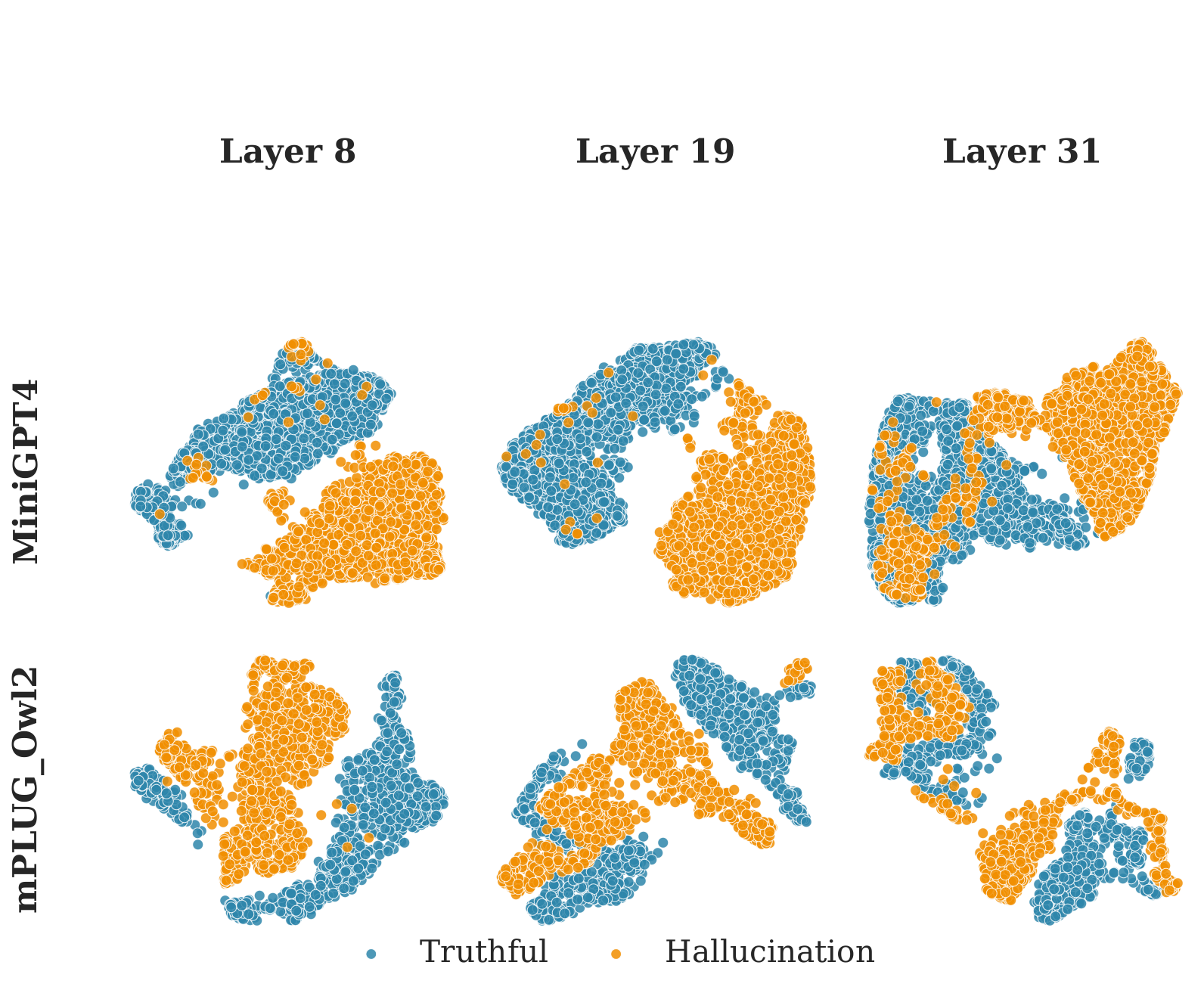}
     \hfill \vspace{-2mm}
    \caption{UMAP visualization of MiniGPT4 and mPLUG-Owl2 hidden activations for truthful (blue) and hallucinatory (amber),  samples from LURE, showing distinct clusters that reveal low-rank hallucination subspaces.}
    \label{fig:tsne}
\end{figure}
\begin{figure*}[t]
\centering
\includegraphics[width=1\textwidth,height=0.35\textwidth]{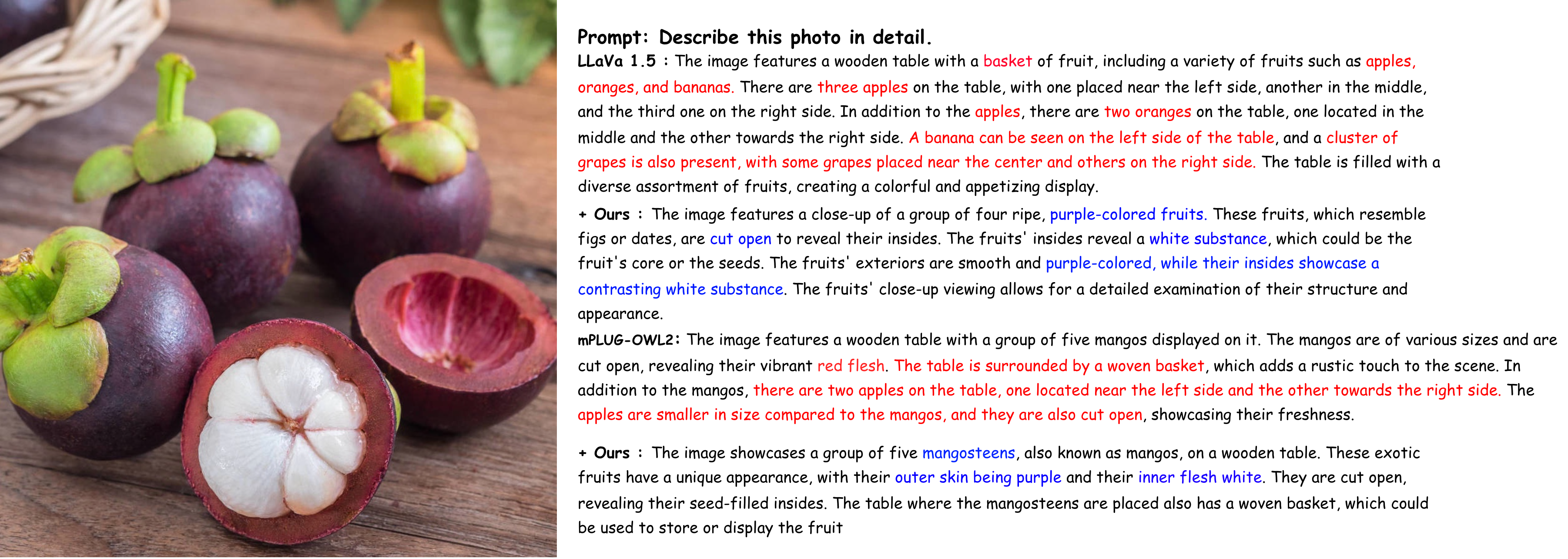}
 \vspace{-4pt}
 \caption{Qualitative comparison of captions generated by LLaVA and mPLUG-Owl2 before and after applying Spectral Representation Filtering (SRF). The baseline model hallucinates non-existent fruits such as apples and bananas, while SRF suppresses these errors.}
 \label{fig:qual_1}
\vspace{-7pt}
\end{figure*}
We address this gap with a gradient-free, inference-efficient correction paradigm that operates at the representational level. Through covariance analysis of hidden states from paired truthful and hallucinatory generations, we identify low-rank hallucination modes, principal directions of variance that encode spurious biases. As visualized in Fig.~\ref{fig:tsne}, UMAP projections of hidden activations reveal that truthful (blue) and hallucinatory (amber) representations, are highly separable in different layers. This structured divergence indicates that hallucinations emerge from localized subspaces within the model’s latent geometry. Leveraging this insight, we apply eigendecomposition of the hallucination covariance to construct a soft spectral filter that selectively damps these modes in the output projections of feed-forward networks within deeper LLM layers. Controlled by a tunable damping factor~$\alpha$, this operation equalizes feature anisotropies without modifying the architecture or requiring retraining, preserving semantic fidelity for truthful generations.

Inspired by spectral methods in signal processing~\cite{oppenheim1999discrete}, our approach incurs zero runtime cost, applying precomputed corrections directly to weights. Evaluations on three VLMs (LLaVA-1.5~\cite{liu2023visual}, MiniGPT-4~\cite{zhu2023minigpt}, mPLUG-Owl2~\cite{ye2023mplugowl}) across MSCOCO captioning, POPE VQA, and visual recognition tasks such as A-OKVQA~\cite{schwenk2022okvqa} and LLaVA-Bench~\cite{liu2024improved}, reveal substantial OH reductions for our method, surpassing prior methods while ensuring broad applicability. A qualitative comparison in Fig.~\ref{fig:qual_1} further illustrates SRF’s effect, showing how the same model, when filtered through our method, eliminates hallucinated objects and yields faithful, visually grounded descriptions.

To summarize, our contributions are:
\textbf{1. A covariance-based dissection of VLM representations}: unveiling structured, low-rank hallucination modes as the geometric signatures of OH.
\textbf{2. Spectral Representation Filtering (SRF)}: a lightweight, plug-in mechanism for soft mode suppression via FFN weight modulation, enabling efficient, generalizable correction. \textbf{3. Comprehensive benchmarking}: establishing SOTA performance in OH mitigation, with ablation insights into design choices for optimal trade-offs.

\section{Related Works}
\textbf{Vision-Language Models}
VLMs combine visual and textual modalities for tasks such as image captioning and VQA. They typically consist of a pre-trained vision encoder (e.g., CLIP-ViT~\cite{radford2021learning}) to extract image features, a projection module that maps these features into the LLM embedding space, and a large language model (e.g., LLaMA~\cite{touvron2023llama} or Vicuna~\cite{chiang2023vicuna}), which is usually frozen during alignment and later instruction-tuned on multimodal datasets.

Modern architectures improve modularity and efficiency through three integration strategies: early fusion~\cite{gao2023llama,zhang2023llamaadapter,liu2023visual,bai2023qwenvlversatilevisionlanguagemodel}, bridging~\cite{dai2023instructblip}, and mid-fusion~\cite{alayrac2022flamingo}. Early fusion prepends projected visual tokens to the LLM input, as in LLaVA~\cite{liu2023visual} and Qwen-VL2~\cite{wang2024qwen2}. Bridging compresses visual tokens into a small set of query embeddings, used in BLIP~\cite{dai2023instructblip} and mPLUG-Owl~\cite{ye2023mplugowl}. Mid-fusion distributes cross-attention across layers, as in Flamingo~\cite{alayrac2022flamingo} and LLaMA 3.2-Vision~\cite{dubey2024llama}, enabling deeper multimodal interaction. \\ \textbf{Object Hallucinations in Vision-Language Models} Visual Hallucinations in vision large language models (Vision LLMs) refer to the generation of plausible but factually incorrect outputs that misrepresent the input visual content~\cite{liu2024survey,li-etal-2023-evaluating,kim2024discovering}, encompassing a range of distortions from subtle inaccuracies to overt fabrications. These include inventing object attributes (e.g., assigning colors, sizes, or states not depicted, such as describing a grayscale image as vibrant), fabricating spatial relationships. In multimodal tasks like visual question answering~\cite{LiaZeh_VisionAmplified_MICCAI2025,10.1145/3664647.3681663,zhu2024combating} and image captioning~\cite{rohrbach-etal-2018-object,biten2022let,zhai2023halle,li-etal-2023-evaluating}, hallucinations typically appear as overly assured claims that merge observed visual elements with embedded textual knowledge, resulting in responses that appear relevant to the prompt on the surface but compromise reliability in scenarios demanding accurate visual interpretation. To mitigate hallucinations in vision large language models (Vision LLMs), methods are broadly classified into training-based~\cite{sun2024aligning,jing2024fgaif,zhao2025looking} and training-free~\cite{huang2024opera,leng2024mitigating,chen2024halc,liu2025reducing,yang2025nullu} categories. Training-based techniques involve retraining components to strengthen vision-language alignment, such as fine-tuning the projection module or LLM with hallucination-aware losses that reward fidelity to input visuals, often leveraging datasets augmented with paired image-text contrasts. Additional strategies include reinforcement learning from human feedback~\cite{sun2024aligning} tailored to multimodal outputs. Recently, FGAIF~\cite{jing2024fgaif} introduces fine-grained AI-generated feedback that provides hallucination supervision (object existence, attributes, and relations), enabling more precise RL-based alignment of LVLMs.  OPERA~\cite{huang2024opera} applies an over-trust penalty and a retrospection-allocation mechanism during decoding to reduce hallucinations. Visual Contrastive Decoding~\cite{leng2024mitigating} contrasts outputs on original versus distorted visuals to suppress object hallucinations. HALC~\cite{chen2024halc} is an adaptive focal-contrast decoding strategy that corrects hallucinated tokens on the fly. Nullu~\cite{yang2025nullu} identifies hallucination-related directions by performing an SVD over feature differences and then removes the top singular vectors via hard projection, fully excising those directions from the model’s representation space. In contrast, our SRF method derives its correction from the eigendecomposition of the hallucination \emph{covariance} a second-order characterization that captures both variance magnitude and cross-dimensional structure. Rather than deleting directions outright, SRF applies smooth spectral damping, yielding a more controlled and stable attenuation that preserves semantic content while suppressing hallucination-dominant modes.

\section{Method}

\subsection{Background and Notation}
\begin{figure*}[t]
\centering
\includegraphics[width=1\textwidth,height=0.4\textwidth]{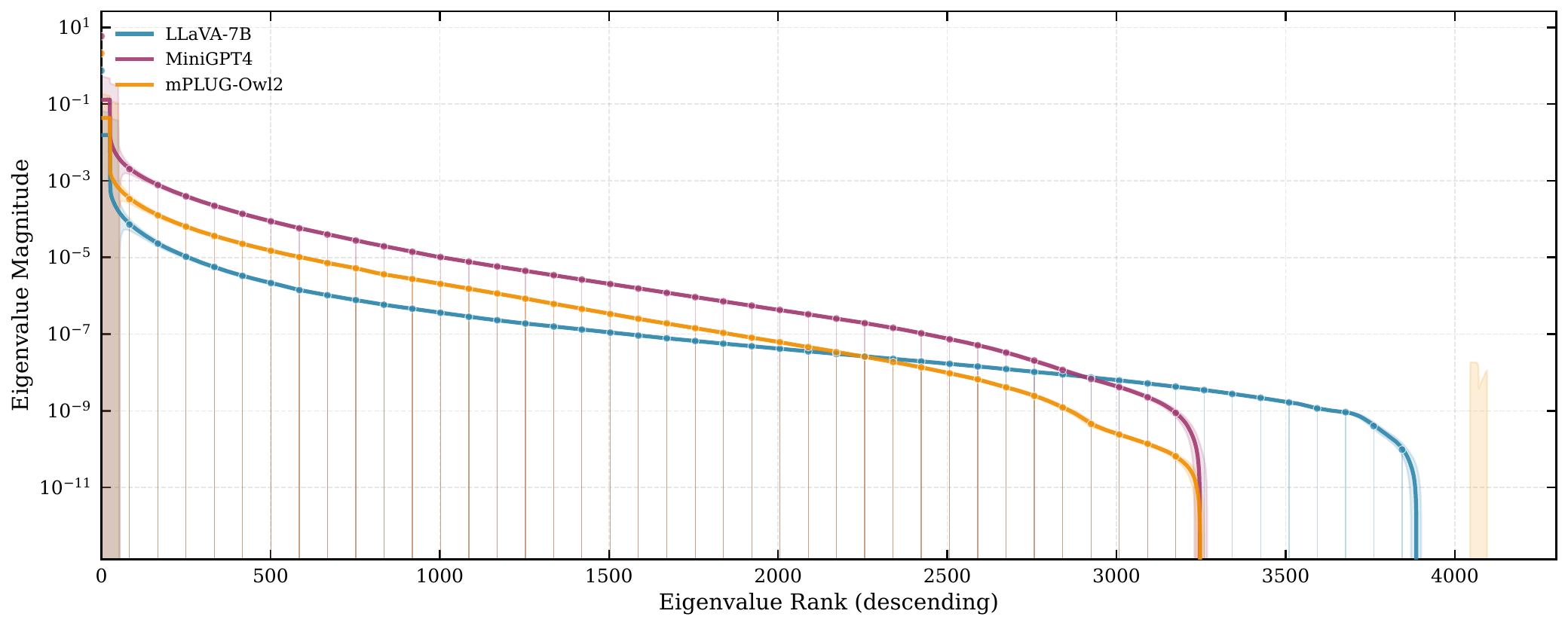}
 \vspace{-4pt}
 \caption{Hallucination spectrum (eigenvalues of $\Sigma_H = Q\Lambda Q^{\top}$) for three 
vision–language models.  
The curves show a small set of high-variance “spikes’’ followed by a long decaying tail, 
indicating that hallucination behavior is concentrated in a low-dimensional set of dominant 
eigenmodes, with the remaining spectrum reflecting noise-like variation.}\label{fig:spectra}
\vspace{-7pt}
\end{figure*}
Most modern vision-language models process multimodal inputs through a three-stage pipeline. First, a vision encoder $\phi_v: \mathcal{I} \rightarrow \mathbb{R}^{n \times d_v}$ converts images into $n$ visual tokens, each with dimension $d_v$. Next, a projection layer $\phi_p: \mathbb{R}^{d_v} \rightarrow \mathbb{R}^{d}$ aligns these tokens with the language model's embedding space of dimension $d$. Finally, a large language model (LLM) with $L$ layers processes the combined visual and textual sequence to generate outputs. At each layer $\ell \in \{1, \ldots, L\}$ of the LLM, hidden states evolve through self-attention and feed-forward blocks. The feed-forward network applies two linear transformations with a non-linearity between them:
\begin{equation}
\text{FFN}_{\ell}(\mathbf{x}) = \mathbf{W}_{\ell}^{\text{out}} \sigma(\mathbf{W}_{\ell}^{\text{in}} \mathbf{x} + \mathbf{b}_{\ell}^{\text{in}}) + \mathbf{b}_{\ell}^{\text{out}},
\end{equation}
where $\mathbf{W}_{\ell}^{\text{out}} \in \mathbb{R}^{d_{\text{ff}} \times d}$ is the output projection from the feed-forward hidden dimension $d_{\text{ff}}$ back to the model dimension $d$, and $\mathbf{W}_{\ell}^{\text{in}} \in \mathbb{R}^{d \times d_{\text{ff}}}$ is the input projection.

For any distribution over feature vectors $\mathbf{x} \in \mathbb{R}^d$, where $d$ is the feature dimension, $\mathbf{\Sigma} = \mathbb{E}[(\mathbf{x} - \boldsymbol{\mu})(\mathbf{x} - \boldsymbol{\mu})^{\top}]$ captures second-order statistics. The eigendecomposition $\mathbf{\Sigma} = \mathbf{Q} \mathbf{\Lambda} \mathbf{Q}^{\top}$ reveals principal directions of variation, where eigenvectors $\{\mathbf{q}_i\}_{i=1}^d$ form an orthonormal basis and eigenvalues $\{\lambda_i\}_{i=1}^d$ measure variance along each direction. A key operation in signal processing is spectral filtering, which transforms a matrix via its eigendecomposition:
\begin{equation}
\mathbf{A}^{(f)} = \mathbf{Q} f(\mathbf{\Lambda}) \mathbf{Q}^{\top},
\end{equation}
where $f: \mathbb{R}_+ \rightarrow \mathbb{R}_+$ operates element-wise.

\subsection{Hallucination Covariance Structure}

Object hallucination occurs when vision–language models describe or refer to objects that do not actually appear in the input image. We hypothesize that such hallucinations emerge from systematic differences in the internal representational geometry of vision–language models. As illustrated in Figure~\ref{fig:tsne}, activations derived from the LURE captioning dataset~\cite{zhou2023analyzing} reveal distinct clustering patterns between truthful and hallucinated responses across layers, suggesting separable manifolds in latent space. To substantiate this hypothesis, we analyze the covariance structure of these activations to characterize layer-wise divergence patterns indicative of hallucination-specific subspaces.

We construct a dataset $\mathcal{D} = \{(I_i, c_i^+, c_i^-)\}_{i=1}^N$ by leveraging the LURE benchmark, where each image $I_i$ is associated with both a hallucinated caption $c_i^+$ generated by the model and a human-verified truthful caption $c_i^-$. For each sample $i$, we extract hidden states from every transformer layer $\ell$ and compute sequence-averaged representations:
\begin{equation} \mathbf{x}_i^+ = \frac{1}{T_i^+} \sum_{t=1}^{T_i^+} \mathbf{h}_{\ell,t}(I_i, c_i^+), \quad \mathbf{x}_i^- = \frac{1}{T_i^-} \sum_{t=1}^{T_i^-} \mathbf{h}_{\ell,t}(I_i, c_i^-), \end{equation}
where $\mathbf{h}_{\ell,t} \in \mathbb{R}^d$ denotes the $d$-dimensional hidden state at layer $\ell$ and token position $t$, and $T_i^+$ and $T_i^-$ are the sequence lengths of the hallucinated and truthful captions, respectively. We then compute the difference vectors:
\begin{equation}
\mathbf{d}_i = \mathbf{x}_i^+ - \mathbf{x}_i^- \in \mathbb{R}^d,
\end{equation}
which quantify representational shifts between hallucinated and truthful responses for the same image.

Aggregating across all $N$ samples, we construct the hallucination covariance matrix:
\begin{equation}
\mathbf{\Sigma}_H = \frac{1}{N} \sum_{i=1}^N \mathbf{d}_i \mathbf{d}_i^{\top} \in \mathbb{R}^{d \times d}.
\end{equation}
This $d \times d$ matrix $\mathbf{\Sigma}_H$ is symmetric and positive semi-definite. It characterizes the second-moment structure of hallucination-induced changes in feature space. Computing the eigendecomposition gives:
\begin{equation}
\mathbf{\Sigma}_H = \mathbf{Q} \mathbf{\Lambda} \mathbf{Q}^{\top} = \sum_{j=1}^d \lambda_j \mathbf{q}_j \mathbf{q}_j^{\top},
\end{equation}
where eigenvalues are ordered $\lambda_1 \geq \lambda_2 \geq \cdots \geq \lambda_d \geq 0$ and $\mathbf{Q} \in \mathbb{R}^{d \times d}$ contains the eigenvectors as columns. The eigenvectors $\mathbf{q}_j \in \mathbb{R}^d$ corresponding to large eigenvalues $\lambda_j$ identify hallucination modes—directions in latent space along which hallucinated and truthful activations diverge most strongly. If hallucinations were merely stochastic noise, the spectrum of $\mathbf{\Sigma}_H$ would be approximately isotropic, i.e., $\mathbf{\Sigma}_H \approx \sigma^2 \mathbf{I}$, yielding a flat distribution.

However, as shown in Figure~\ref{fig:spectra}, all models exhibit highly anisotropic spectra with a few dominant eigenvalues spanning several orders of magnitude. This concentration of variance indicates that hallucinations occupy low-dimensional, structured subspaces rather than emerging from random perturbations. Such dominant modes likely correspond to semantic priors or overgeneralized object templates internalized during vision–language pretraining, reflecting the model’s tendency to project uncertain visual evidence onto frequently co-occurring linguistic concepts.

\subsection{Soft Suppression via Spectral Filtering}

Instead of completely removing hallucination modes through hard projection, we use soft spectral filtering to smoothly reduce their influence. We define a suppression operator:
\begin{equation}
\mathbf{P}_{\alpha} = \mathbf{Q} f_{\alpha}(\mathbf{\Lambda}) \mathbf{Q}^{\top} \in \mathbb{R}^{d \times d},
\label{eq:spf}
\end{equation}
where the damping function is:
\begin{equation}
f_{\alpha}(\lambda_j) = \frac{1}{1 + \alpha \lambda_j}, \quad \alpha > 0.
\end{equation}
This function attenuates high-variance modes: when $\alpha \lambda_j$ is large, the mode is strongly suppressed with $f_{\alpha}(\lambda_j) \approx 1/(\alpha \lambda_j)$, while when $\alpha \lambda_j$ is small, the mode is mostly preserved with $f_{\alpha}(\lambda_j) \approx 1 - \alpha \lambda_j$. The scalar parameter $\alpha$ controls the strength of suppression, providing smooth interpolation between no correction ($\alpha \to 0$ giving $\mathbf{P}_{\alpha} \to \mathbf{I}$) and complete suppression ($\alpha \to \infty$ giving $\mathbf{P}_{\alpha} \to \mathbf{0}$).

The $d \times d$ operator $\mathbf{P}_{\alpha}$ acts as a variance equalizer. Hallucinations create excess variance along certain directions, making the feature distribution non-uniform. By applying $f_{\alpha}(\lambda_j)$, we reduce variance along high-variance directions, moving the hallucination covariance toward a more uniform structure, obtaining the transformed covariance:
\begin{equation}
\mathbf{P}_{\alpha} \mathbf{\Sigma}_H \mathbf{P}_{\alpha} = \mathbf{Q} \left(\frac{\mathbf{\Lambda}}{(1 + \alpha \mathbf{\Lambda})^2}\right) \mathbf{Q}^{\top} \in \mathbb{R}^{d \times d},
\end{equation}
showing reduced variance along all directions, with stronger reduction for larger eigenvalues.

\subsection{Weight Correction and Inference}

We apply the suppression operator to the feed-forward weights in selected layers. Specifically, for each layer $\ell$ in a chosen set $\mathcal{L}$—corresponding in our experiments to the deeper layers where semantic abstraction and language grounding predominantly occur (see Section~\ref{sec:implementation_details})—we attenuate the subspace directions associated with dominant hallucination modes. We compute corrected weights:
\begin{equation}
\mathbf{W}_{\ell}^{\text{corr}} = \mathbf{P}_{\alpha} \mathbf{W}_{\ell}^{\text{out}} \in \mathbb{R}^{d_{\text{ff}} \times d}.
\end{equation}
This modulates the transformation from the $d_{\text{ff}}$-dimensional feed-forward hidden states to the $d$-dimensional output representations, suppressing components aligned with hallucination modes. Since $\mathbf{P}_{\alpha}$ is precomputed and integrated into the weights, this adds no extra cost during inference and the corrected model runs at the same speed as the original.

The complete algorithm proceeds as follows. For each layer $\ell \in \mathcal{L}$, we first extract features $\mathbf{X}^+ = [\mathbf{x}_1^+, \ldots, \mathbf{x}_N^+] \in \mathbb{R}^{d \times N}$ and $\mathbf{X}^- = [\mathbf{x}_1^-, \ldots, \mathbf{x}_N^-] \in \mathbb{R}^{d \times N}$ from the contrastive dataset, each column represents a $d$-dimensional feature vector for one of the $N$ samples. We compute difference vectors $\mathbf{d}_i = \mathbf{x}_i^+ - \mathbf{x}_i^- \in \mathbb{R}^d$ for $i = 1, \ldots, N$ and form the hallucination covariance $\mathbf{\Sigma}_H = \frac{1}{N} \sum_{i=1}^N \mathbf{d}_i \mathbf{d}_i^{\top} \in \mathbb{R}^{d \times d}$. We then perform eigendecomposition $\mathbf{\Sigma}_H = \mathbf{Q} \mathbf{\Lambda} \mathbf{Q}^{\top}$ and construct the filter matrix $f_{\alpha}(\mathbf{\Lambda}) = \text{diag}(\frac{1}{1 + \alpha \lambda_1}, \ldots, \frac{1}{1 + \alpha \lambda_d}) \in \mathbb{R}^{d \times d}$, which is a diagonal matrix. The suppression operator is $\mathbf{P}_{\alpha} = \mathbf{Q} f_{\alpha}(\mathbf{\Lambda}) \mathbf{Q}^{\top} \in \mathbb{R}^{d \times d}$, which is used to compute corrected weights $\mathbf{W}_{\ell}^{\text{corr}} = \mathbf{P}_{\alpha} \mathbf{W}_{\ell}^{\text{out}} \in \mathbb{R}^{d_{\text{ff}} \times d}$. Finally, we update the model by replacing the weights with the corrected ones.
The method has two main parameters. First, the layer set $\mathcal{L} \subseteq \{1, \ldots, L\}$ determines which of the $L$ total layers to correct. We apply corrections to deeper layers where semantic processing happens, typically using $\mathcal{L} = \{16, 17, \ldots, 32\}$ for 32-layer models. Second, the damping parameter $\alpha$ controls the trade-off between reducing hallucinations and keeping semantic information. 

Our approach differs from projection-based methods, including but not limited to~\cite{yang2025nullu}, which use operators like $\mathbf{I} - \mathbf{V}_k \mathbf{V}_k^{\top} \in \mathbb{R}^{d \times d}$ to completely remove the top $k$ modes, where $\mathbf{V}_k \in \mathbb{R}^{d \times k}$ contains the first $k$ eigenvectors. Hard projection can cause loss of semantic information, while our soft suppression keeps all directions but reduces problematic ones gradually. This provides smoother control and better preserves features while still reducing hallucinations.
\section{Experiments}
\begin{figure}[t]
    \centering
     \hfill
    \includegraphics[width=1\linewidth,height=0.5\linewidth]{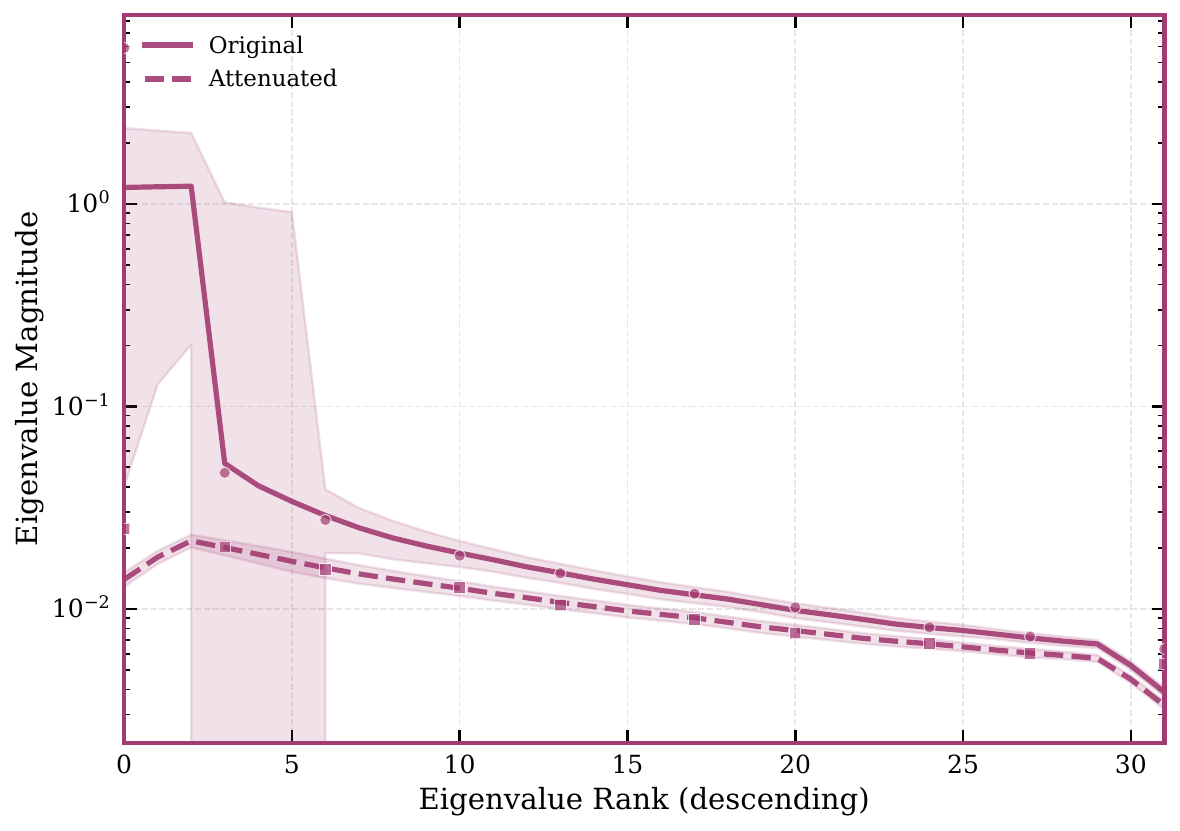}
     \hfill \vspace{-2mm}
    \caption{Effect of the spectral suppression operator on the MiniGPT-4 hallucination spectrum. 
The attenuated curve (dashed) selectively contracts dominant hallucination-aligned 
eigenmodes while leaving lower-variance structure largely unchanged.
}
    \label{fig:eigen}
\end{figure}
\subsection{Spectral Suppression and Selection of \texorpdfstring{$\alpha$}{alpha}}
\label{ssec:alpha_selection}

We select the suppression parameter~$\alpha$ using a spectrum-driven rule grounded in the
geometry of the hallucination covariance~$\Sigma_H$. The \emph{Soft Spectral Suppression}
operator (Eq.~\ref{eq:spf}) contracts variance along hallucination-aligned eigenmodes, with
$\alpha$ governing the overall attenuation strength. As illustrated in
Fig.~\ref{fig:spectra}, the spectrum displays a sharply peaked leading component followed by
a flat tail, enabling a stable, model-agnostic criterion for choosing~$\alpha$.

\myparagraph{Matheuristic Selection of $\alpha$.}
We determine $\alpha$ by specifying the desired contraction of the dominant eigenvalue
$\lambda_1$. Controlling the suppression ratio
\begin{equation}
    \rho_1(\alpha) = \frac{1}{1 + \alpha\lambda_1}
\end{equation}
gives rise to the closed-form rule
\begin{equation}
    \alpha = \frac{1 - \eta}{\eta\,\lambda_1},
\end{equation}
where $\eta$ represents the retained fraction of variance permitted in the
most hallucination-aligned direction. In all experiments, we set $\eta = 0.1$. This
procedure ties $\alpha$ directly to the strength of the leading eigenmode, while
lower-variance directions remain largely unaffected due to their substantially smaller
eigenvalues (Fig.~\ref{fig:eigen}).

\myparagraph{Interpretation.}
Because $\alpha$ scales proportionally with the dominant spectral magnitude of~$\Sigma_H$,
the resulting attenuation is both stable and model-adaptive. Crucially, $\alpha$ is not
chosen using downstream metrics; improvements arise naturally from aligning the suppression
strength with the spectral structure itself, yielding a practical operating range on the
order of \(1\text{--}10^{2}\), as detailed in Section~\ref{ssec:sensitivity_alpha}.

\begin{figure*}[t]
\centering
\includegraphics[width=1\textwidth,height=0.4\textwidth]{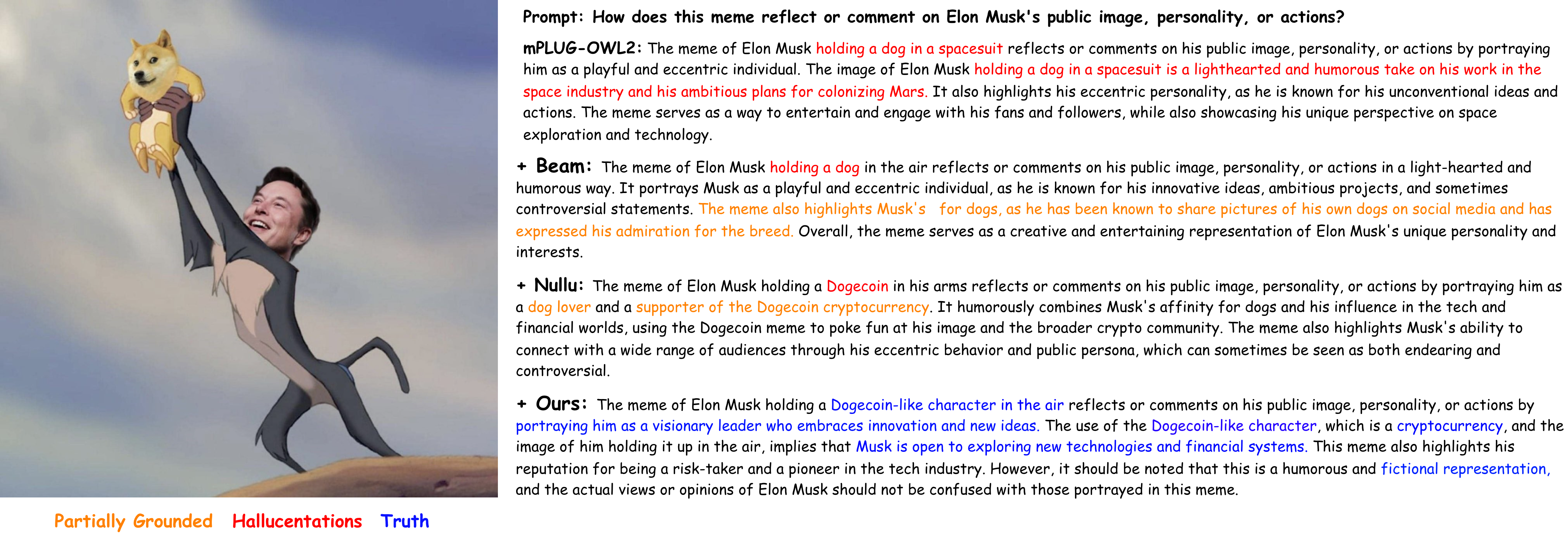}
 \vspace{-4pt}
 \caption{A qualitative case study comparing model-generated descriptions for a complex meme. We visualize the outputs from the Greedy Baseline, BEAM, Nullu, and our method, color-coding text segments as \textcolor{red}{Hallucinations}, \textcolor{orange}{Partially Grounded}, or \textcolor{blue}{Truth}.}\label{fig:qual_results}
\vspace{-7pt}
\end{figure*}

\subsection{Dataset \& Models and Baselines}
\textbf{Evaluation Datasets} We assess the efficacy of SRF across four complementary hallucination-centric benchmarks, each probing distinct failure modes in vision-language generation. (1) CHAIR (Caption Hallucination Assessment with Image Relevance)~\cite{rohrbach-etal-2018-object} evaluates hallucination frequency by comparing generated captions with MSCOCO~\cite{lin2014microsoft} object annotations. It reports $\text{CHAIR}_S$ (sentence-level rate: captions with spurious objects) and $\text{CHAIR}_I$ (object-level rate: fraction of hallucinated objects). Lower scores indicate better faithfulness, using 500 MSCOCO validation images and averaging results over three random seeds. (2)POPE (Poll-based Object Probing Evaluation) [20] is a VQA-style binary benchmark where models answer yes/no questions about object presence in images. It includes three negative sampling strategies: random (uniform object selection), popular (high-frequency but absent objects to expose bias), and adversarial (semantically related objects for harder grounding). We report accuracy, precision, and F1 scores on 500 MSCOCO images with ten queries each. (3) A-OKVQA~\cite{schwenk2022okvqa} extends OK-VQA with more diverse and challenging reasoning tasks. It contains ~25k questions with multiple-choice and free-form answers, requiring commonsense, visual, factual, and physical reasoning. Unlike simple lookup tasks, A-OKVQA demands compositional reasoning over the visual scene. (4) To further examine the generalization of SRF beyond object-level hallucination metrics, we evaluate it on the LLaVA-Bench~\cite{liu2024improved} benchmark. LLaVA-Bench consists of 24 high-quality images, each accompanied by a detailed human-written description and a curated set of questions designed to probe multi-turn reasoning, visual grounding, and linguistic coherence. Following the evaluation setup in~\cite{yang2025nullu}, we prompt GPT-4 to act as an automatic evaluator, comparing the baseline and edited model responses. The judge assigns two independent scores, one for \emph{accuracy}, reflecting the absence of hallucinated or inconsistent content, and one for \emph{detailedness}, reflecting the richness and completeness of non-hallucinatory visual details. Both scores are given on a 1–10 scale.

\textbf{Vision-Language Models.}
We evaluate SRF on three open-source VLMs representing different vision-language fusion paradigms. LLaVA-1.5~\cite{liu2023visual} uses early fusion, concatenating CLIP~\cite{radford2021learning} visual tokens with text embeddings for processing by Vicuna-7B, emphasizing efficiency and modularity. MiniGPT-4~\cite{zhu2023minigpt} applies bridging with Q-Former–compressed visual features injected into LLaMA-7B, balancing token reduction and representation. mPLUG-Owl2~\cite{ye2023mplugowl2} follows mid-fusion with modular connectors and dual-branch pathways for understanding and generation. 

\textbf{Baselines} We benchmark SRF against  hallucination-mitigation methods. Decoding approaches intervene at inference without changing weights: Beam Search~\cite{freitag2017beam} ranks multiple candidate sequences to reduce greedy artifacts, while Visual Contrastive Decoding (VCD)~\cite{leng2024mitigating} biases logits away from high-variance tokens. Weight-editing methods modify model parameters offline, enabling inference-time corrections with no extra computation. Nullu~\cite{yang2025nullu} removes hallucination by projecting weights orthogonally to low-rank subspaces identified via eigendecomposition of feature differences, fully eliminating top-$k$ hallucination modes. SRF, in contrast, applies soft spectral attenuation with $f_\alpha(\lambda) = \frac{1}{1 + \alpha\lambda}$, reducing hallucination-correlated modes proportionally to their eigenvalues while preserving all directions. This approach allows finer control over correction strength, avoids abrupt feature disruption, and maintains similar computational efficiency.
\begin{table*}[ht]
\small
\centering
\setlength{\tabcolsep}{2pt}
{
\begin{tabular}{ l | c c | c c | c c}
\toprule
{\textbf{Method}} 
&\multicolumn{2}{c|}{LLaVA-1.5}
&\multicolumn{2}{c|}{MiniGPT-4}
&\multicolumn{2}{c}{mPLUG-Owl2} \\
&\textbf{CHAIR}$_S \downarrow$ &\textbf{CHAIR}$_I \downarrow$ &\textbf{CHAIR}$_S \downarrow$ &\textbf{CHAIR}$_I \downarrow$ &\textbf{CHAIR}$_S \downarrow$ &\textbf{CHAIR}$_I \downarrow$  \\ 
\midrule
Greedy
& $19.46_{\pm 1.50}$ &  $6.26_{\pm 0.11}$& $28.06_{\pm 0.60}$ & $11.24_{\pm 0.09}$& $20.60_{\pm 2.20}$& $6.66_{\pm 0.57}$ \\
Beam Search \cite{freitag2017beam}
& $18.26_{\pm 0.73}$ &$5.89_{\pm 0.29}$ & $25.26_{\pm 0.89}$& $10.76_{\pm 0.55}$  & $18.53_{\pm 1.97}$& $6.03_{\pm 0.60}$  \\
VCD \cite{leng2024mitigating}
& $21.40_{\pm 1.48}$& $7.43_{\pm 0.52}$& $26.82_{\pm 0.31}$ & $10.02_{\pm 0.01}$ & $17.52_{\pm 0.78}$ & $5.98_{\pm 0.22}$ \\ 
Nullu \cite{chen2024halc} w/o Beam
& $19.99_{\pm 1.40}$ &$6.68_{\pm 0.56}$ &$25.66_{\pm 1.93}$& $11.02_{\pm 0.65}$& $16.39_{\pm 0.65}$ & $5.82_{\pm 0.14}$    \\ 
Nullu \cite{chen2024halc} w Beam
& $18.80_{\pm 0.43}$& $6.51_{\pm 0.24}$& $22.66_{\pm 0.67}$ & $9.79_{\pm 0.39}$ & $17.06_{\pm 1.08}$& $5.93_{\pm 0.32}$ \\ 
\midrule
\rowcolor{mygray} \textbf{\nameshort} 
& $15.46_{\pm 1.60}$& $5.42_{\pm 0.57}$  & $17.80_{\pm 1.81}$& $8.80_{\pm 0.68}$  & $\textbf{15.40}_{\pm \textbf{1.01}}$ & $\textbf{5.34}_{\pm \textbf{0.46}}$ 
\\ 
\rowcolor{mygray} \textbf{\nameshortbeam} 
& $\textbf{14.73}_{\pm \textbf{1.65}}$ & $\textbf{5.18}_{\pm \textbf{0.45}}$ & $\textbf{16.46}_{\pm \textbf{0.49}}$& $\textbf{8.35}_{\pm \textbf{0.64}}$ & $\textbf{15.40}_{\pm \textbf{1.57}}$ & $5.41_{\pm 0.54}$
\\
\bottomrule
\end{tabular}}
\caption{Comparison of LVLMs (LLaVA-1.5, MiniGPT-4, and mPLUG-Owl2) on the MSCOCO dataset using various strategies to reduce object hallucination. Metrics CHAIRS and CHAIRI measure the amount of hallucination, with lower values indicating fewer errors. Experiments were performed with a maximum token limit of 64.}
\label{tab:coco_results}
\end{table*}
\subsection{Implementation Details}
\label{sec:implementation_details}
All experiments are conducted using the official models implementations, with our post-hoc corrections applied to the LLM's deeper layers (layers 16--32), all of our corrections are applied only over the down\_proj layers in Feed Forward blocks. Inference uses greedy decoding unless specified, with a maximum token limit of 64 for CHAIR and POPE, 1 for A-OKVQA and 1024 for Llava-Bench. For the Nullu~\cite{yang2025nullu} benchmark, we ran multiple configurations and report results using the best-performing one. In all of the CHAIR and LLaVA-Bench evaluations we use $\alpha = 70$, while for POPE and AOKVQA we set $\alpha=6$.
\begin{table*}[htp]
\small
\centering
\resizebox{0.8\textwidth}{!}{
    \begin{tabular}{llccc|ccc|ccc}
    \toprule
    {\textbf{Setting}} & \textbf{Method} & \multicolumn{3}{c}{\textbf{LLaVA-1.5}} & \multicolumn{3}{c}{\textbf{mPLUG-Owl2}} & \multicolumn{3}{c}{\textbf{MiniGPT4}} \\
    \cmidrule(lr){3-5} \cmidrule(lr){6-8} \cmidrule(lr){9-11}
    & & \textbf{Acc} & \textbf{Prec} & \textbf{F$_1$} & \textbf{Acc} & \textbf{Prec} & \textbf{F$_1$} & \textbf{Acc} & \textbf{Prec} & \textbf{F$_1$} \\
    \midrule
    \multirow{4}{*}{\textit{Random}} 
          & Greedy & 88.33 & 86.81 & 88.57 & 82.53 & 76.12 & 84.44 & 53.60 & 51.92 & 67.69 \\
          & Nullu & 88.93 & 89.89 & 88.80 & 84.27 & 79.34 & 85.49 & 57.80 & 54.37 & 69.70 \\
          & Ours & \textbf{89.20} & \textbf{90.61} & \textbf{89.01} & \textbf{87.20} & \textbf{85.14} & \textbf{87.56} & \textbf{58.13} & \textbf{54.59} & \textbf{69.81} \\
    \midrule
    \multirow{4}{*}{\textit{Popular}} 
          & Greedy & 83.93 & 81.07 & 84.64 & 71.33 & 64.65 & 76.66 & 50.40 & 50.21 & 66.12 \\
          & Nullu & 85.60 & 84.68 & 85.79 & 73.60 & 67.02 & 77.88 & 52.27 & 51.18 & 67.25 \\
          & Ours & \textbf{85.67} & \textbf{85.06} & \textbf{85.80} & \textbf{79.53} & \textbf{74.21} & \textbf{81.56} & \textbf{52.73} & \textbf{51.44} & \textbf{67.33} \\
    \midrule
    \multirow{4}{*}{\textit{Adversarial}} 
          & Greedy & 74.00 & 68.75 & 77.19 & 66.73 & 61.08 & 73.50 & 50.07 & 50.03 & 66.03 \\
          & Nullu & 75.20 & 70.72 & \textbf{77.62} & 69.33 & 63.63 & 74.64 & 51.40 & 50.73 & 66.73 \\
          & Ours & \textbf{75.53} & \textbf{71.21} & 77.52 & \textbf{73.80} & \textbf{68.65} & \textbf{76.98} & \textbf{52.12} & \textbf{51.12} & \textbf{67.01} \\
    \bottomrule
    \end{tabular}%
    }
\caption{POPE results with random, popular and adversarial samplings compared to existing OH mitigation methods.}
\label{tab:pope}
\end{table*}
\subsection{Qualitative Analysis}
Figure \ref{fig:qual_results} shows a qualitative example using a meme of Elon Musk in The Lion King. The \textbf{Greedy} and \textbf{BEAM} models hallucinate heavily—introducing nonexistent objects or misidentifying the character—while \textbf{Nullu} adds strong inferential bias, confidently framing the scene as related to Dogecoin. In contrast, our method stays grounded in the image, identifying only a Dogecoin-like figure and describing the scene as a playful parody. This example illustrates how our correction reduces speculative associations and anchors responses more firmly in visual evidence.
\subsection{Quantitative Results}
\myparagraph{CHAIR} Table~\ref{tab:coco_results} reports hallucination suppression and caption quality on the CHAIR benchmark across three representative VLM architectures. Our Spectral Representation Filtering (SRF) achieves consistent reductions in both sentence-level and object-level hallucination rates across all models. For LLaVA-1.5, SRF without beam search (\nameshort)) lowers $\text{CHAIR}_S$ from $19.46{\pm1.50}$ to $15.46_{\pm1.60}$ and $\text{CHAIR}_I$ from $6.26{\pm0.11}$ to $5.42_{\pm0.57}$. The beam-augmented variant (\nameshortbeam)) further improves $\text{CHAIR}_S$ to $14.73{\pm1.65}$ and $\text{CHAIR}_I$ to $5.18\pm 0.45$. Similar behavior is observed for MiniGPT-4, where $\text{CHAIR}_S$ and $\text{CHAIR}_I$ decrease from $28.06{\pm0.60}$ and $11.24{\pm0.09}$ under greedy to $16.46{\pm 0.49}$ and $8.35{\pm 0.64}$ respectively. For mPLUG-Owl2, SRF attains $15.40{\pm1.01}$ and $5.34{\pm0.46}$ on $\text{CHAIR}_S$ and $\text{CHAIR}_I$, outperforming both Nullu and decoding-time baselines such as VCD and Beam Search. These consistent improvements demonstrate that soft spectral filtering effectively suppresses spurious object mentions while preserving linguistic fluency.

\myparagraph{POPE} The POPE evaluation in Table~\ref{tab:pope} assesses factual grounding across random, popular, and adversarial object sampling regimes. SRF achieves the highest F$_1$ scores across nearly all model and setting combinations, reflecting improved visual grounding and reduced false positives.
For LLaVA-1.5, SRF reaches $89.01$ F$_1$ under the random split and $85.80$ under popular split, slightly exceeding Nullu and decoding baselines. Under the most challenging adversarial setting, SRF maintains an F$_1$ of $77.52$, compared to $77.19$ for the greedy baseline and $77.62$ for Nullu.
In mPLUG-Owl2, SRF consistently improves factual accuracy across all baselines in all of the splits, with F$_1$ increasing from $84.44$ to $87.56$ in the random split, $76.66$ to $81.56$ in the popular split, and from $73.50$ to $76.98$ in the adversarial split. MiniGPT-4 exhibits similar trends, where SRF marginally improves grounding performance in all settings, maintaining the highest precision and F$_1$ across the board.
These results indicate that spectral suppression of high-variance hallucination directions improves consistency across diverse setups, reinforcing SRF’s robustness across VLM architectures.

\myparagraph{Llava-Bench} Results are presented in Table ~\ref{tab:ablations_llava_bench}, our method achieves the highest scores across both models, with MiniGPT-4 reaching 4.95 in accuracy and 4.82 in detailness, and mPLUG-OWL2 reaching 5.50 and 4.71, respectively. Baseline strategies such as Greedy and Beam perform noticeably lower, with accuracy ranging from 4.35 to 5.41 and detailness from 3.58 to 4.18. The Nullu~\cite{yang2025nullu} approach improves over these baselines in some cases, particularly increasing MiniGPT-4’s accuracy to 4.73 and mPLUG-OWL2’s detailness to 4.66, but our method surpasses it on all metrics, showing clear overall improvements in both correctness and response richness.
These results show that SRF is robust: all nonzero $\alpha$ values improve hallucination metrics on CHAIR, and a broad interval of $\alpha$ yields strong suppression without compromising fluency. This behavior is consistent with theoretical expectations for attenuating anisotropic covariance modes and validates the stability of SRF under varying damping strengths.
\subsection{Sensitivity of the Suppression Parameter \texorpdfstring{$\alpha$}{alpha}}
\label{ssec:sensitivity_alpha}

We analyze the robustness of the spectral suppression mechanism by varying the
damping parameter~$\alpha$ over a wide range while keeping all other components
fixed. Since $\alpha=0$ corresponds to the unmodified baseline, increasing
$\alpha$ progressively strengthens the attenuation of hallucination-dominant
eigenmodes according to the spectral scaling rule in
Sec.~\ref{ssec:alpha_selection}.

\begin{figure}[t]
    \centering
     \hfill
    \includegraphics[width=1\linewidth]{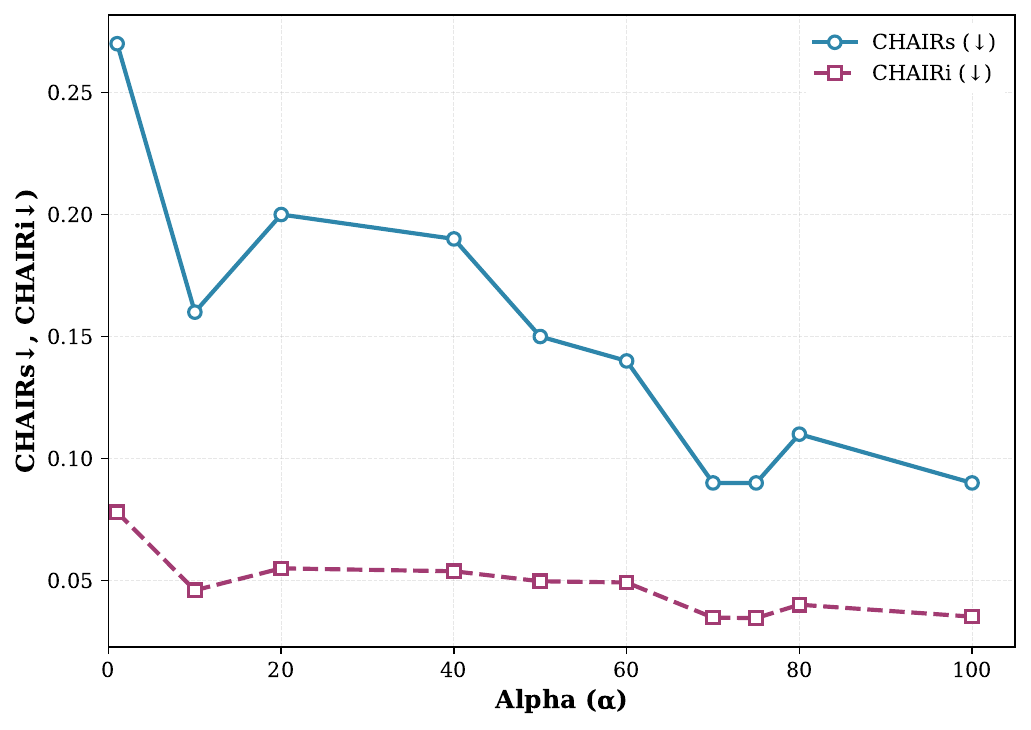}
     \hfill \vspace{-2mm}
    \caption{\textbf{Sensitivity of SRF to the damping parameter $\alpha$ on a random subset of 100 sampels of the CHAIR validation set for LLaVA-1.5-7B.}
  $\alpha = 0$ corresponds to the unmodified baseline model.}
    \label{fig:sensitivty}
\end{figure}

Figure~\ref{fig:sensitivty} shows the sensitivity behaviour across
$\alpha\in[1,100]$ on 100 randomly sampled CHAIR instances. We observe a stable
and monotonic reduction in hallucination-related activations as $\alpha$
increases, with improvements persisting across a broad interval rather than a
single finely tuned value. This reflects the structure of the hallucination
spectrum, where the leading eigenvalues dominate the suppression dynamics while
lower-variance modes remain largely unaffected.

Although $\alpha=70$ is used in our main experiments for strong attenuation of
the dominant modes, the sensitivity curve indicates that many values in the
upper range yield comparable trade-offs, demonstrating that the method operates
within a wide range of effective suppression and improvement over the baseline (\(\alpha=0\)).

\subsection{Ablation Studies}
We conduct a series of ablation experiments on the A-OKVQA benchmark to analyze the sensitivity of our Spectral Representation Filtering (SRF) framework to different design components. The results, including the most prominent baselines, are depicted in Table~\ref{tab:ablations}. Unless otherwise stated, we use the default configuration with $\alpha=6$ and intervention layers $\ell \in$ $[16-32]$.

\myparagraph{Effect of Mean Subtraction} To evaluate whether our gains stem merely from centering the representational space, we replace the spectral projection operator $P_{\alpha}$ with a simple mean subtraction of the hallucination direction $\mu_d$ from each representation. As shown in the first row of Table~\ref{tab:ablations}, this baseline yields a marginal improvement over the base model, confirming that removing the dominant offset alone is insufficient. The structured projection offered by $P_{\alpha}$ captures higher-order variance directions essential for mitigating hallucination artifacts.

\myparagraph{Hard vs.\ Soft Projection.}  
We compare our soft attenuation operator \(P_{\alpha} = Q\,f_{\alpha}(\Lambda)\,Q^{\top}\), which scales the contribution of each eigenmode by \(f_{\alpha}(\lambda_j)=1/(1+\alpha\lambda_j)\), to a conventional hard projection that fully removes the top-\(k\) eigenvectors via \((I - V_k V_k^\top)\). While hard projection suppresses the largest hallucination modes more aggressively, it also discards potentially useful semantic directions and leads to degraded performance. In contrast, our smooth attenuation retains all directions but re-weights them according to their variance magnitude, offering a finer-grained trade-off between factual grounding and representation quality. The ablation results show that the soft approach not only reduces hallucination more effectively, but also better preserves semantic coherence compared to hard removal.

\myparagraph{Using SVD Instead of Covariance-Based Spectra.}
We additionally evaluate a variant of SRF in which the hallucination covariance 
$\Sigma_H$ is replaced by the singular vectors obtained from an SVD decomposition 
of the feature-difference matrix. This substitutes second-order structure with a 
purely first-order factorization. As shown in Table~\ref{tab:ablations}, the SVD 
variant provides a noticeable improvement over the baseline (75.28 vs.\ 72.49), 
but remains below the full SRF configuration (75.46). This performance gap 
indicates that the covariance spectrum which captures both variance magnitudes 
and cross-dimensional interactions is essential for reliably identifying 
hallucination-aligned directions. In contrast, SVD alone lacks the anisotropic 
second-order structure required for effective suppression, confirming the 
importance of covariance geometry in our method.
\begin{table}[t]
\small
\centering
    \begin{tabular}{@{}l@{~}c@{~}c@{~}c@{~}c@{~}c@{}}
    \toprule
    {\textbf{Ablation}} & \multicolumn{2}{c}{\textbf{MiniGPT4}} & \multicolumn{2}{c}{\textbf{mPLUG-OWL2}}\\ 
    \cmidrule(lr){2-3} \cmidrule(lr){4-5}
    & {\textbf{Accuracy}} $\uparrow$ & {\textbf{Detailness}} $\uparrow$ & {\textbf{Accuracy}} $\uparrow$ & {\textbf{Detailness}} $\uparrow$\\
    \midrule
    Greedy & 4.35 & 3.58 & 5.05 & 4.18\\
    Beam & 4.38 & 4.06 & 5.41 & 4.16\\
    Nullu & 4.73 & 4.13 & 5.16 & 4.66\\
    Ours & \textbf{4.95} & \textbf{4.82}& \textbf{5.50} & \textbf{4.71}\\
    \midrule
    \end{tabular}%
\caption{GPT4-As-a-Judge evaluation over the LLaVA-Bench dataset, the numbers are on scale of 10.}
\label{tab:ablations_llava_bench}
\end{table}
\begin{table}[t]
\small
\centering
    \begin{tabular}{ll@{~~~~~~~~~~~~~~~}ll}
\toprule
\textbf{Method} & \textbf{Acc} &
\textbf{Method} & \textbf{Acc} \\
\cmidrule(lr){1-2}
\cmidrule(r){3-4}
Greedy & 72.49 &
$\ell \in [16\!-\!24]$ & 72.58 \\
VCD~\cite{leng2024mitigating} & 73.92 &
$\ell \in [24\!-\!32]$ & 72.49 \\
Nullu~\cite{yang2025nullu} & 75.02 &
$\ell \in [8\!-\!24]$ & 72.66 \\
SVD & 75.28 &
 $\ell \in [8\!-\!16]$ & 72.84  \\
$\mu_d$ Substraction & 72.75 &
Ours Full & \textbf{75.46} \\
\midrule
\end{tabular}
\caption{\textbf{Results (including ablations) on A-OKVQA with LLaVA-1.5-7B (\texorpdfstring{$\alpha=6.0$}{alpha=6.0}).}
Different intervention strategies and layer ranges are compared, showing that suppression applied to deeper layers yields the strongest gains, with the full SRF configuration achieving the best overall performance.}
\label{tab:ablations}
\end{table}

\myparagraph{Layer-wise Intervention.}  
To determine the most effective representational locus for correction, we apply SRF at different layers. Early-layer interventions have limited impact, as low-level features do not  encode semantic hallucination modes. In contrast, mid-to-late layers show consistent improvements, with the best performance achieved for $\ell \in [16,32]$, This indicates that hallucination directions emerge predominantly in high-level semantic spaces.\section{Conclusions}
This work addresses object hallucinations in vision-language models by analyzing the covariance structure of hidden states and introduces Spectral Representation Filtering (SRF), a training-free method that suppresses hallucination-dominant eigenmodes through soft spectral attenuation of feed-forward weights.

SRF offers three key advantages: it operates entirely post-hoc without architectural modifications, incurs zero inference overhead by precomputing weight corrections, and provides smooth control over hallucination suppression through soft damping rather than hard projection. Comprehensive evaluations across three VLM families (LLaVA-1.5, MiniGPT-4, mPLUG-Owl2) and four benchmarks (CHAIR, POPE, A-OKVQA, LLaVA-Bench) demonstrate state-of-the-art hallucination reduction while maintaining linguistic fluency and visual reasoning capabilities. By treating hallucinations as structured covariance anomalies amenable to spectral correction, SRF establishes a principled, efficient pathway for enhancing the reliability of multimodal AI systems.

\bibliography{main}
\bibliographystyle{abbrv}
\appendix
\onecolumn

\end{document}